\documentclass[10pt,twocolumn,letterpaper]{article}

\usepackage{iccv}
\usepackage{times}
\usepackage{epsfig}
\usepackage{graphicx}
\usepackage{amsmath}
\usepackage{amssymb}
\usepackage{multirow}
\usepackage{bm}
\usepackage{setspace}
\usepackage{threeparttable}

\usepackage[pagebackref=true,breaklinks=true,letterpaper=true,colorlinks,bookmarks=false]{hyperref}

\iccvfinalcopy 

\def\iccvPaperID{8} 

\ificcvfinal\fi
\begin{document}

\title{Facial Pose Estimation by Deep Learning
from Label Distributions}

\author{Zhaoxiang Liu\\
		CloudMinds\\
		{\tt\small robin.liu@cloudminds.com}
		\and
	    Zezhou Chen\\
		CloudMinds\\
		{\tt\small chenzezhou007@aliyun.com}
	    \and
		Jinqiang Bai\\
		Beihang University\\
		{\tt\small baijinqiang@buaa.edu.cn}
        \and
		Shaohua Li\\
		CloudMinds\\
		{\tt\small shaohua.li@cloudminds.com}
        \and
		Shiguo Lian\\
		CloudMinds\\
		{\tt\small sg\_lian@163.com}
}

\maketitle

\begin{abstract}
  Facial pose estimation has gained a lot of attentions in many practical applications, such as human-robot interaction, gaze estimation and driver monitoring. Meanwhile, end-to-end deep learning-based facial pose estimation is becoming more and more popular. However, facial pose estimation suffers from a key challenge: the lack of sufficient training data for many poses, especially for large poses.  Inspired by the observation that the faces under close poses look similar, we reformulate the facial pose estimation as a label distribution learning problem, considering each face image as an example associated with a Gaussian label distribution rather than a single label, and construct a convolutional neural network which is trained with a multi-loss function on AFLW dataset and 300W-LP dataset to predict the facial poses directly from color image. Extensive experiments are conducted on several popular benchmarks, including AFLW2000, BIWI, AFLW and AFW, where our approach shows a significant advantage over other state-of-the-art methods.
\end{abstract}

\section{Introduction}\label{sec:intro}

Facial pose estimation has received more and more attentions in the past few years \cite{1,2,3,4,5,6,7,8,9,10,11,12,17,18,19}, it plays an important role in many practical applications such as driver monitoring \cite{1,2}, human-robot or human-computer interaction \cite{3,4,5,6,DBLP:journals/corr/abs-1908.07262,DBLP:journals/corr/abs-1908-06607,DBLP:journals/corr/abs-1908.07750}, gaze estimation \cite{7,8,9,10}, human behavior analysis \cite{11}, face alignment \cite{12,13} and face recognition \cite{14}. All of these unconstrained scenarios require a facial pose estimator which is resistant to environmental variations (\eg occlusion, pose, illumination and resolution variations).

Though some good results have been made by using commercial depth cameras \cite{15}, one limitation that could not be neglected lies in that depth camera does not work well under uncontrolled environment where sunlight or ambient light is strong, and it often needs more space and more power compared to monocular RGB camera. These impede its feasibility in real-world applications \cite{16,17}.

Traditionally, facial pose can be computed by estimating some facial key-points from target face and solving 2D to 3D correspondence with a mean 3D head model. Though facial key-point estimation has been recently improved greatly by deep learning \cite{18}, facial pose estimation is inherently a two-step process which is error-prone. The accuracy of the pose estimate depends upon the quality of key-points as well as the 3D head model. If the localized key-points are inaccurate or inadequate,the estimate of pose becomes poor or the pose estimation may even become infeasible. Additionally, generic 3D head models can also bring in errors for any given individual, and deforming the head model to adapt to each individual demands significant amounts of data and computation.

\begin{figure*}[t]
	\centering
	\begin{tabular}{ccc}
		\includegraphics[width=0.3\linewidth]{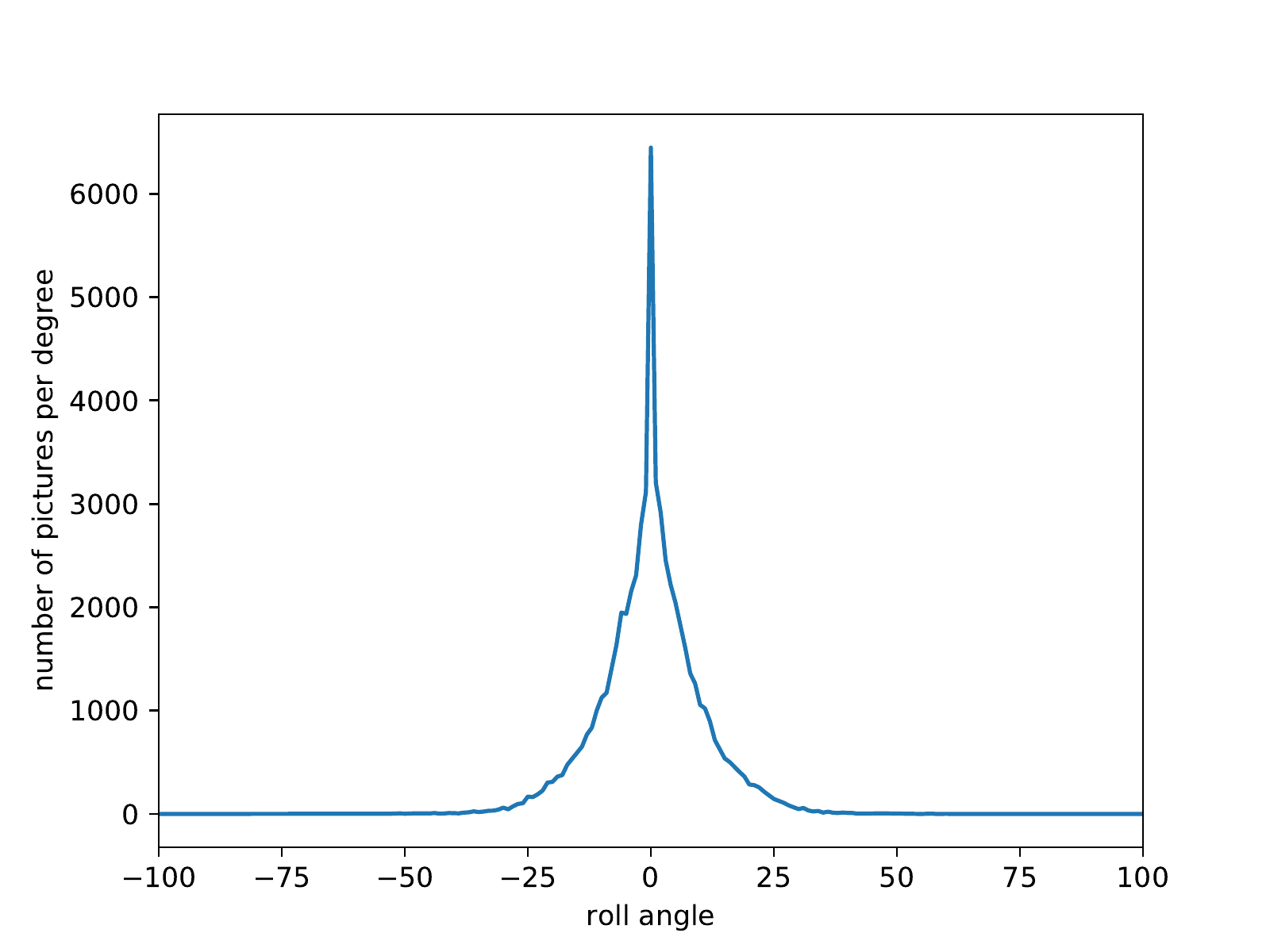} &
		\includegraphics[width=0.3\linewidth]{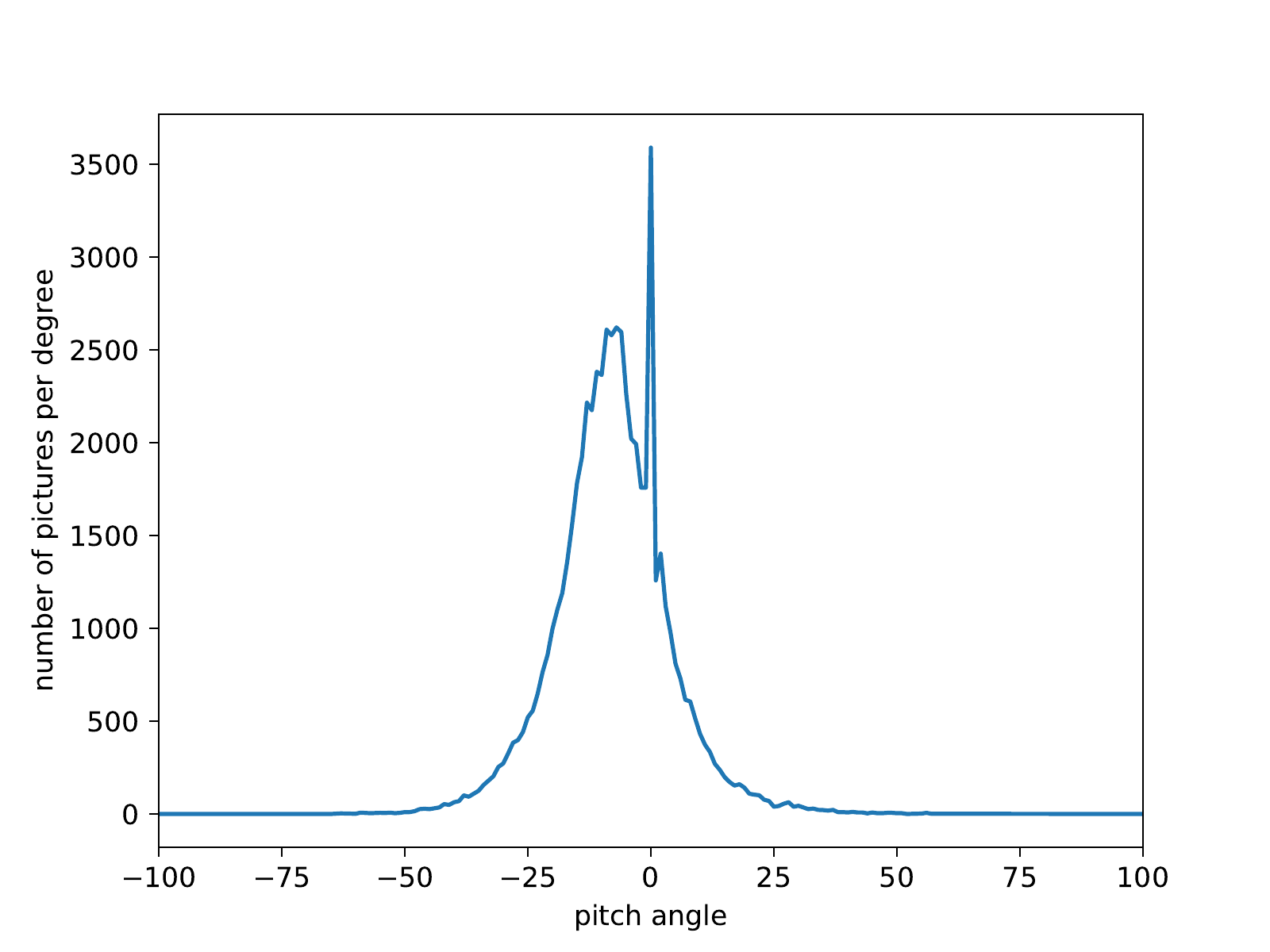} &
		\includegraphics[width=0.3\linewidth]{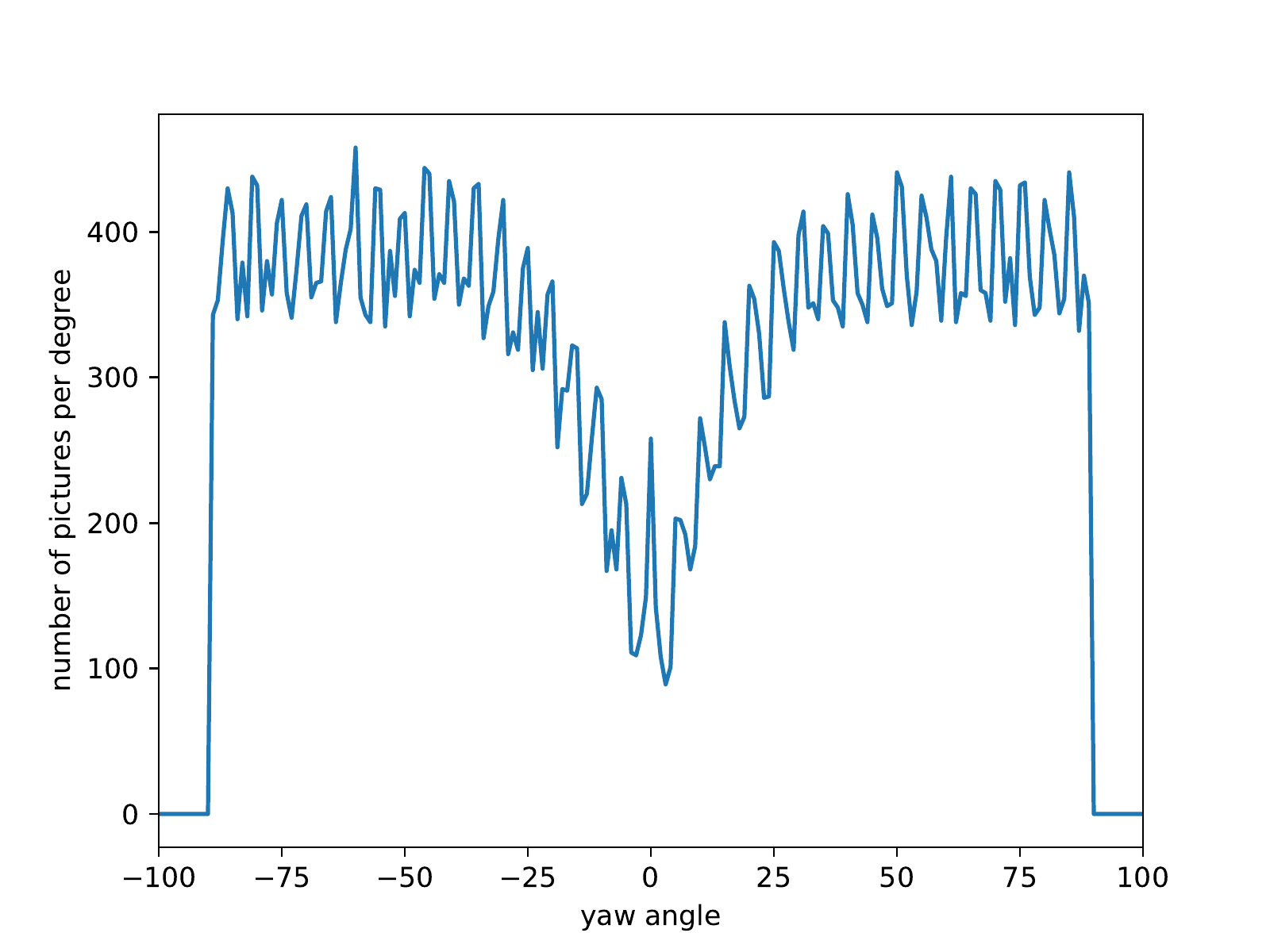}\\
		\includegraphics[width=0.3\linewidth]{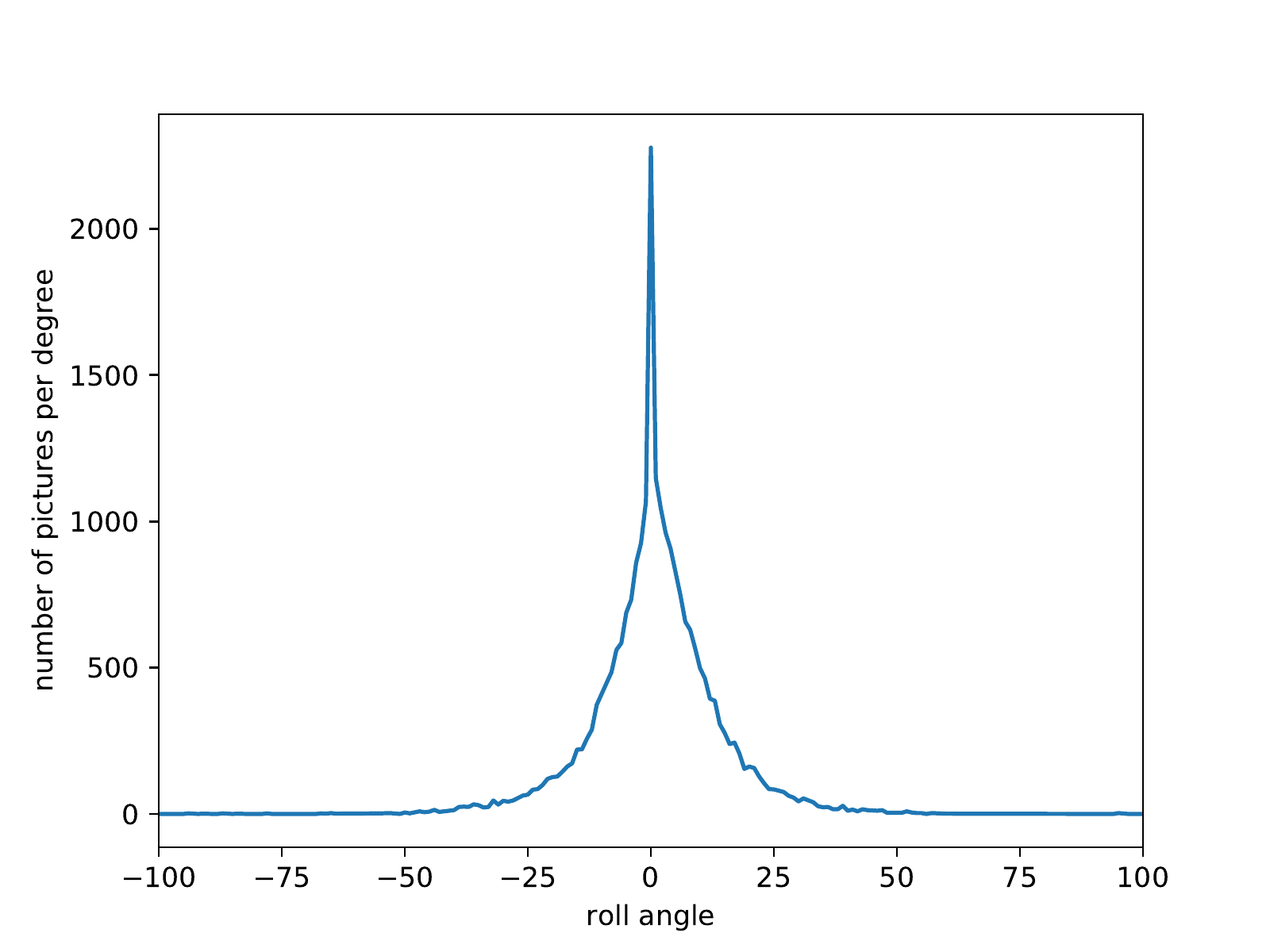} &
		\includegraphics[width=0.3\linewidth]{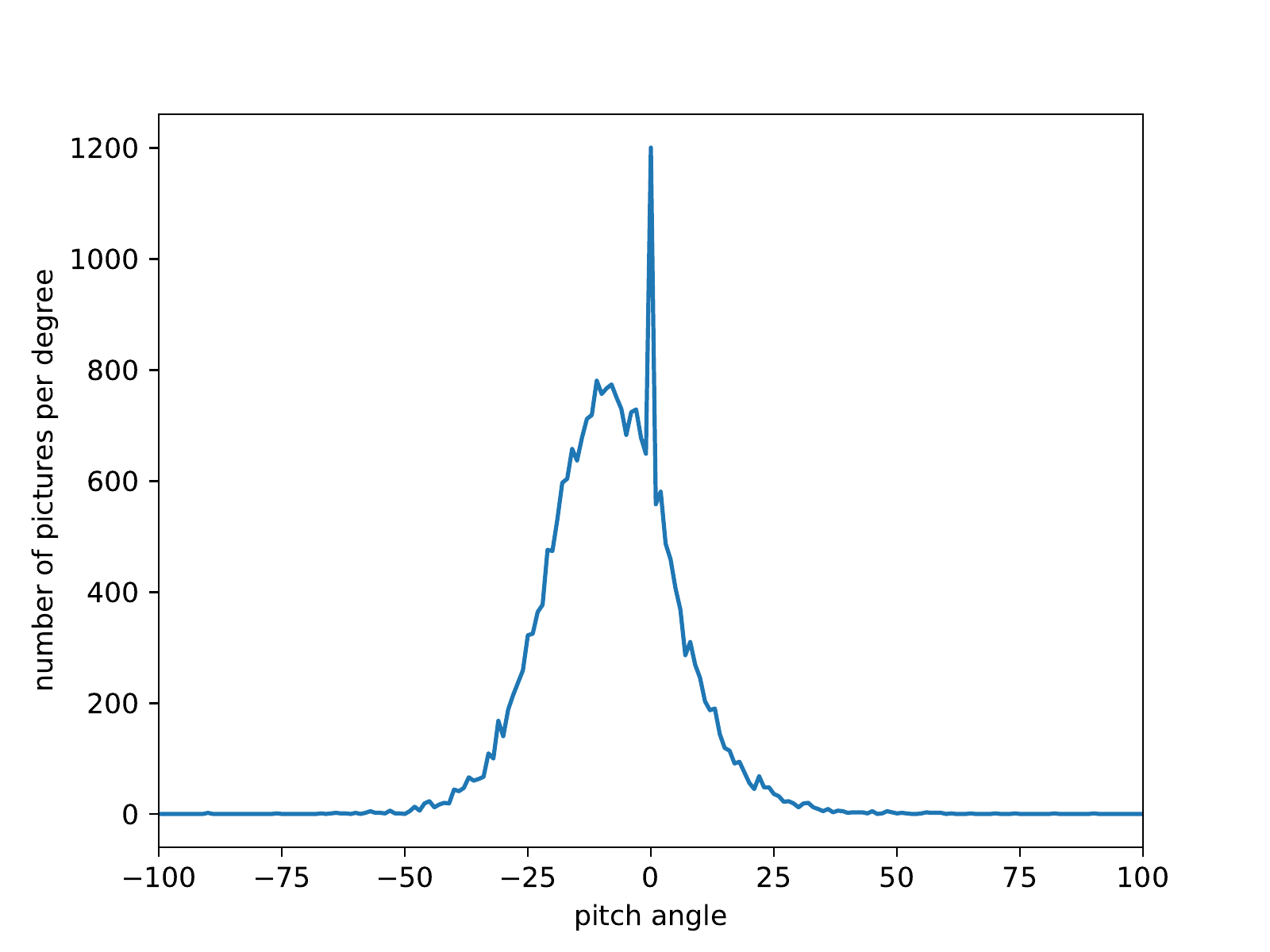} &
		\includegraphics[width=0.3\linewidth]{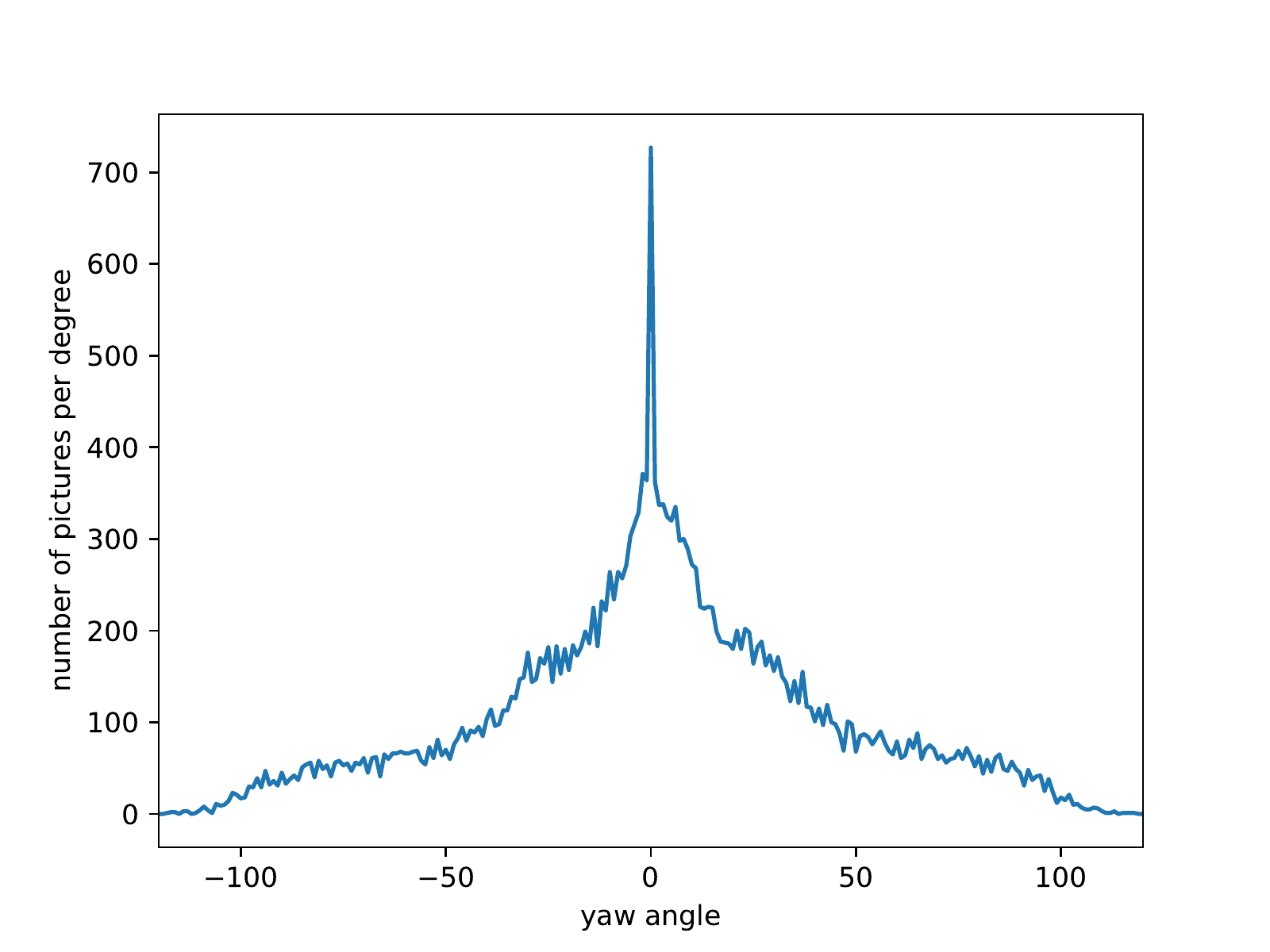}\\
	\end{tabular}
	\caption{The sample distributions of two popular datasets. 300W-LP \cite{53} (first row), AFLW \cite{54} (second row). We can see that most faces lie in the area of small poses.}
	\label{fig:1}
\end{figure*}

Recently, it has become more popular to estimate facial pose end to end using deep learning due to its robustness to environmental variations. The deep learning-based methods have large advantages compared to traditional landmark-to-pose methods, for they always output a pose prediction which does not rely on landmark detection and 3D head model. However, the deep learning-based facial pose estimation has not been thoroughly investigated overall. In some of these cases, facial pose estimation is just one branch of multi-tasks for face analysis, which is used to improve the performances of these other tasks (\eg face detection, key-points localization and gender recognition). The facial pose branch was not designed dedicatedly in terms of accuracy. Some other deep learning-based methods have dedicatedly addressed the facial pose estimation as a pose regression from image \cite{17,19,20,21} using convolution neural networks(CNN), while the work in \cite{16} has concluded that the combination of binned classification and regression works better than regression solely. However, all these deep learning-based methods ignore an important fact that the distribution of training samples is quite imbalanced and there are not sufficient training samples for the large poses. To varying degrees, the most popular datasets for facial pose estimation, such as AFLW \cite{54} and 300W-LP \cite{53}, exist this problem (as shown in Fig.\ref{fig:1}) which can degrade the accuracy of pose estimate, especially for large pose. We argue that it is unreasonable to use soft-max cross-entropy loss for facial pose estimation when training samples are considerably imbalanced, and the accuracy of facial pose estimate still has potential to be improved furtherly. For these losses ignore the similarity between adjacent poses, not taking the relationship between adjacent poses into consideration, other appropriate constraint should be introduced into the loss function.

To this end, we reformulate the facial pose estimation as label distribution learning problem and introduce a more intuitive similarity constraint: Gaussian label distribution loss into the training for facial pose estimation to improve the accuracy. The main contributions of our work can be summarized as follows:
\begin{itemize}
	\item We reveal the fact that the lack of sufficient straining samples exists in the popular facial pose datasets. And we explain why it is not optimal to use soft-max cross-entropy loss for facial pose estimation under this situation. 
	\item We introduce a novel Gaussian label distribution loss into the training for facial pose estimation,  the Gaussian label distribution loss which constrains the similarities between neighbouring poses and can effectively mitigate the insufficiency of training samples, and dramatically boost the accuracy of facial pose estimate.
	\item We demonstrate the effectiveness of our method in facial pose estimation by various comparative experiments. Trained on publicly available datasets, such as AFLW \cite{54}dataset and 300W-LP \cite{53} dataset, our method achieves the-state-of-art results on AFLW2000 \cite{53}, BIWI \cite{52}, AFLW \cite{54} and AFW \cite{55} benchmarks.
\end{itemize}
\section{Related Works}
So far a variety of efforts on facial pose estimation have been dedicated. All these methods can be easily divided depending on whether they use 2D camera or depth camera. Since our work is concerned with deep learning-based method using RGB image from a monocular camera, any other methods using the depth camera will not be considered here. A more detailed description of depth camera-based methods can be found in a recent survey \cite{22} and other previous works \cite{7,15,23,24,25}.

Some early classic studies \cite{26,27,28} can be categorized as appearance template methods which match a view of a person's face to a set of exemplars with corresponding pose labels in order to find the most similar view. For example, the method in \cite{28} adopts support vector machine (SVM) to model the appearance of human faces across multiple views and performs pose estimation by using nearest-neighbor matching. However, the appearance template methods suffer from some limitations. They can only estimate discrete pose without the use of some interpolation method, and they also suffer from the accuracy concerns when the facial region is not localized accurately and efficiency concerns when the exemplar set is very large.

The face detector arrays \cite{29,30} whose idea is to train multiple face detectors for different facial poses once became popular as the success of frontal face detection \cite{31,32,33}. The method in \cite{30} uses a sequence of five multi-view face detectors to estimate facial pose. It is evident that many face detectors are required for each corresponding discrete facial pose, and it is difficult to implement a real time facial pose estimator with a large detector array.

Facial pose estimation can also be formulated as a manifold embedding problem that the high dimensional face image can be embedded into a low dimensional manifold in which facial pose is estimated. Any dimensionality reduction technique can be considered as a part of manifold embedding category. The methods in \cite{34,35} project a face image into a PCA or KPCA subspace and in which compare the result to a set of embedded templates. The method in \cite{36} uses Isometric Feature Mapping (Isomap) to embed a face image into a nonlinear manifold which represents the pose-varying faces. These approaches ignore the pose labels that are available during training and operate in an unsupervised fashion. This results in that the built manifolds not only describe the pose variations but also identity variations \cite{43}. The method \cite{37} utilizes the feature correspondence of identity-invariant geometric features to learn a similarity kernel that only reflects the pose variation ignoring other sources of variation. This method shows a good reliability on benchmark dataset. However, further research is still needed to achieve state-of-the-art performance.

Facial pose estimation can be naturally formulated as a nonlinear regression problem which learns a nonlinear mapping from images to poses. The methods in \cite{38,39,40} adopt support vector regressor(SVR) to estimate the facial pose after a series of preprocessing, including face region cropping, Sobel filtering, PCA \cite{38}, priori knowledge-based linear projection \cite{39}, or localized gradient orientation histogram \cite{40}. The methods in \cite{41,42} utilize multilayer perception(MLP) to regress the facial pose. These methods have one disadvantage that they are prone to error from poor face localization. Recently thank to the great success of deep learning techniques, it has become popular to estimate facial pose using CNN which is robust to shift, scale and distortion. The method in \cite{17} presents an in-depth study of CNN trained on AFLW dataset using L2 regression loss and tested on the Prima, AFLW and AFLW datasets. The method in \cite{19} proposes a GoogLeNet-based architecture trained on AFLW dataset which can predict the key-points and facial pose jointly and reports the pose results on AFLW dataset and AFW \cite{55}dataset. L2 Euclidean loss function is adopted to train the pose predictor which is used to improve key-point localization. The method in \cite{12} also trains a pose estimator using 300W dataset to assist face alignment. Both the method in \cite{20} and the method in \cite{21} build a multitask learning framework for face analysis, including face detection, face alignment, face recognition, pose estimation, age prediction, gender recognition and smile detection. Both methods utilize AFLW dataset to train pose regressors and pose results are also reported on AFLW dataset and AFW dataset. The method in \cite{16} makes an extensive study of combination of classification loss and regression loss on benchmark datasets, including 300w-LP dataset, AFLW dataset, BIWI \cite{52} dataset and AFW dataset, and concludes that the combination of binned classification and regression works better than regression solely. However, all these deep learning-based methods pay no attention to the lack of sufficient training data for many poses. Consequently, the performance of facial pose estimator is limited. This reason motivates us to seek a better solution in this paper.

The label distribution learning (LDL) is a novel machine learning paradigm recently proposed for facial age estimation \cite{44,45}. The LDL is based on the observation that age is ambiguous and faces with adjacent ages are strongly correlated. The main idea of LDL is to utilize adjacent ages when learning a particular age. And a label distribution covers a number of class labels, representing the degree that each label describes the instance. Hence, the LDL is able to deal with insufficient and incomplete training data. Some other problems which share the same characteristic as facial age estimation, such as facial attractiveness computation \cite{46}, crowd counting \cite{47} and pre-release movie rating prediction \cite{48} have achieved outstanding performances by using LDL.

\begin{figure*}[htbp]
	\centering
	\begin{tabular}{cccc}
		\includegraphics[width=0.15\linewidth]{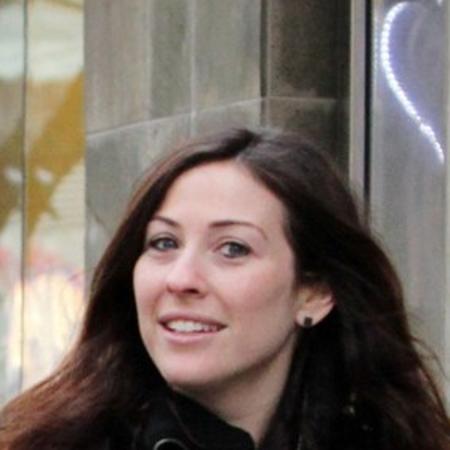} &
		\includegraphics[width=0.15\linewidth]{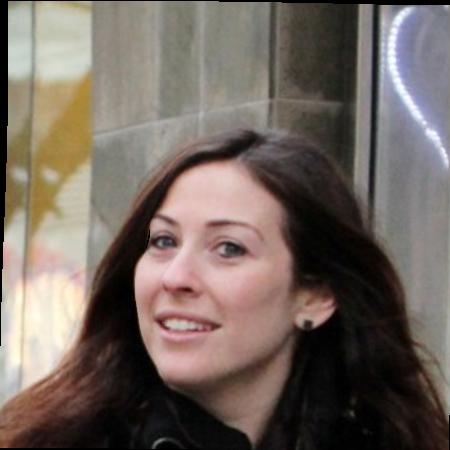} &
		\includegraphics[width=0.15\linewidth]{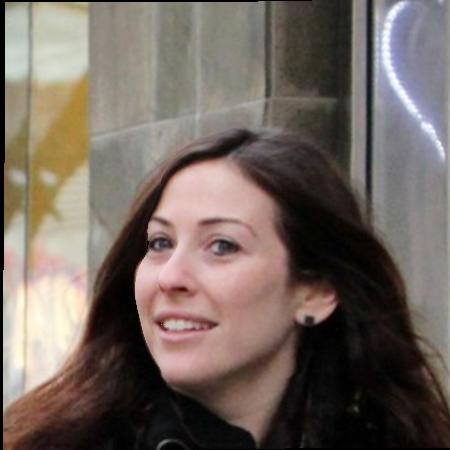} &
		\includegraphics[width=0.15\linewidth]{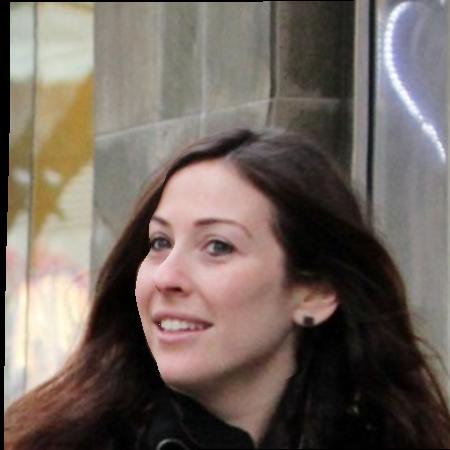} \\
		(a) $yaw=24.8^\circ$ & (b) $yaw=29.8^\circ$ & (c) $yaw=34.8^\circ$ & (d) $yaw=39.8^\circ$ \\
	\end{tabular}
	\caption{The faces of one subject under different poses. }
	\label{fig:2}
\end{figure*}

Facial pose appears similar to facial age, \ie the faces under close poses look similar (as shown in Fig.\ref{fig:2}), the changing of facial pose can be regarded as a relative slow and smooth process and faces under adjacent poses are highly correlated. Thus, the LDL paradigm is an ideal match for the task of facial pose estimation. We notice that similar learning paradigms\cite{geng2014head,gao2017deep} have been proposed to mitigate label ambiguity in head pose estimation. However, they only focused on 2D head pose estimation and were not extensively investigated on such precisely annotated benchmarks as AFLW2000 and BIWI.

\section{Method}
\subsection{Gaussian Label Distribution Learning}
We argue that the lack of sufficient training samples can degrade the accuracy of pose estimator. The reason is that the soft-max cross-entropy loss function used in training encodes the distance between all poses equally and does not take the relationship between adjacent poses into consideration. So it cannot effectively handle the insufficiency problem of training samples. Inspired by the previous work on age estimation \cite{44,45} and facial attractiveness ranking \cite{46}, we reformulate the facial pose estimation as a label distribution learning problem.

It is apparent that the faces under close poses look quite similar (as shown in Fig.\ref{fig:2}). Consequently, additional knowledge about the faces with different poses can be introduced to reinforce the learning problem. It is straightforward to utilize faces under neighboring poses while learning a particular pose. To achieve this, we assign a label distribution to each face image rather than a single label of real pose. This can make a face image contribute to not only the learning of its real pose, but also the learning of its neighbouring poses. We employ three Gaussian label distributions to describe a face example in the yaw, pitch and roll domain respectively to reinforce the whole learning process.

\begin{figure}[htbp]
	\centering
	\includegraphics[scale=0.3]{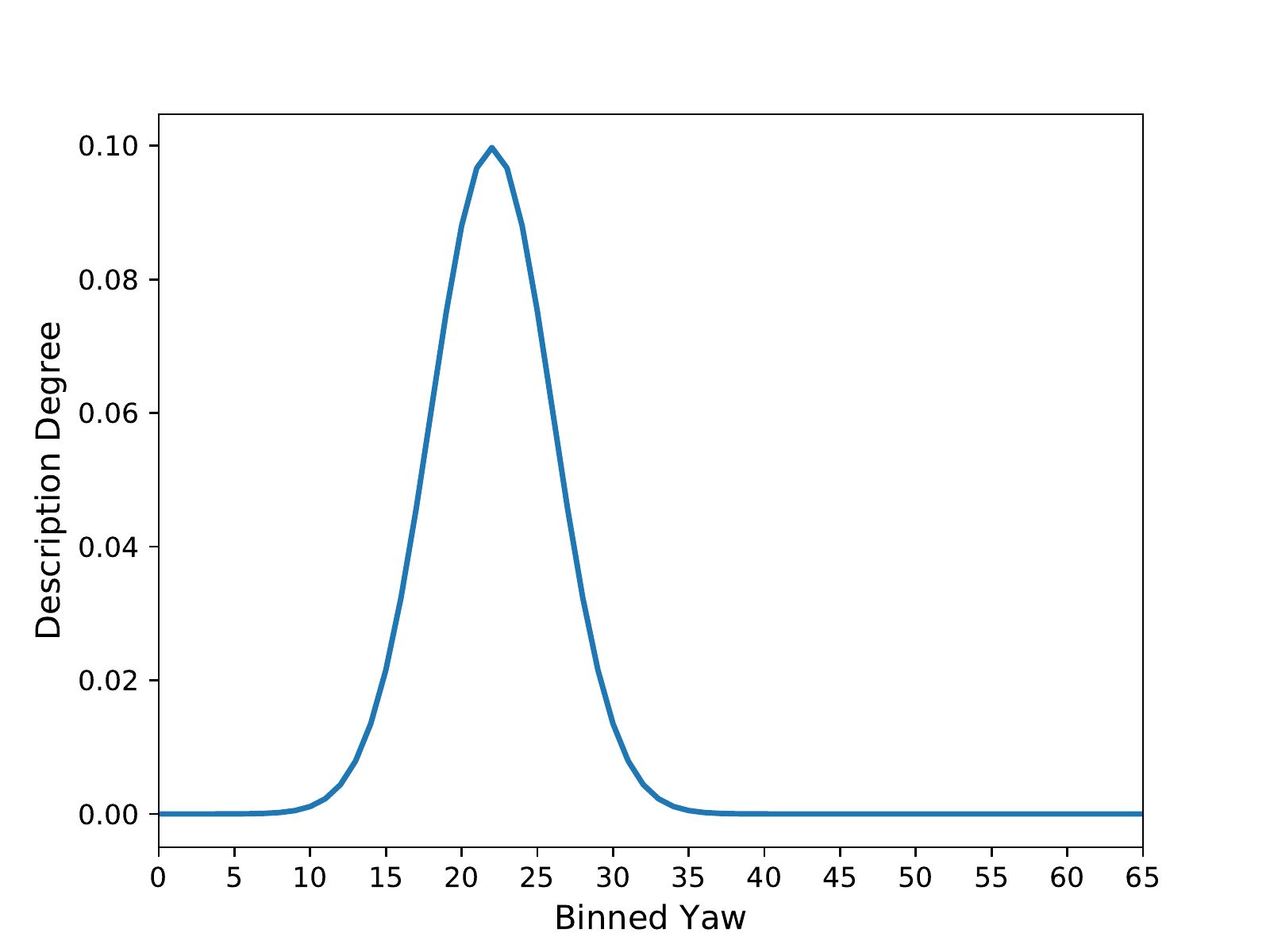}
	\caption{Gaussian label distribution with $\sigma_y=4$ for the ground-truth $yaw=-30^\circ$.}
	\label{fig:4}
\end{figure}

Here we take the yaw as an example to illustrate the Gaussian label distribution. Given a face image $x_i$ and a complete set of yaw labels $\bm y=\{y_1,y_2,\dots,y_M\}$, if its yaw label is $y_\alpha, \alpha=1,2,\dots,M$, then the corresponding yaw label distribution is represented as a multi-dimension vector $\bm D_i^y=\{d_{x_i}^{y_1},d_{x_i}^{y_2},\dots,d_{x_i}^{y_M}\}$, with the $l$-th dimension as follows:
\begin{equation}
\begin{array}{l}
d_{x_i}^{{y_l}} = \frac{{\overline {d_{x_i}^{{y_l}}} }}{{\mathop \sum \nolimits_{u = 1}^M  {\overline {d_{x_i}^{{y_u}}} }}},\\
\overline {d_{x_i}^{{y_l}}} = exp(\frac{{ - {{(l - \alpha )}^2}}}{{2{{\sigma^2}_{y}}}})/{\sigma _{{y}}}, l=1,2,\dots,M
\end{array}
\end{equation}
where $l$ denotes the $l$-th binned yaw, $\alpha$ is the binned ground-truth yaw, $\sigma_y$ is the label standard deviation, and $M$ is the dimension of the yaw label vector which also implicitly represents the maximum yaw. Consequently, $d_{x_i}^{y_l}$ represents the degree that the label $y_l$ describes the example $x_i$ under the constraint $\sum \nolimits_{l = 1}^M {d_{x_i}^{y_l}}=1$, meaning that the label set $\bm y$ fully describes the example. Fig.\ref{fig:4} demonstrates an example of Gaussian label distribution for yaw.

Following the same definition, another two label distributions: $\bm D_i^p=\{ d_{x_i}^{p_1},d_{x_i}^{p_2},\dots,d_{x_i}^{p_N} \}$ and $\bm D_i^r=\{d_{x_i}^{r_1},d_{x_i}^{r_2},\dots,d_{x_i}^{r_K}\}$ can be obtained for $x_i$ with a set of pitch labels $ \bm p=\{p_1,p_2,\dots,p_N\}$ and a set of roll labels $\bm r=\{r_1,r_2,\dots,r_k\}$ respectively as follows:
\begin{equation}
\begin{array}{l}
d_{x_i}^{p_j}=\frac{\overline{d_{x_i}^{p_j}}}{\sum \nolimits_{v = 1}^N {\overline {d_{x_i}^{{p_v}}} }},\\
\overline {d_{x_i}^{{p_j}}} = exp(\frac{{ - {{(j - \beta )}^2}}}{{2{{\sigma^2} _{p}}}})/{\sigma _{{p}}}, j=1,2,\dots,N
\end{array}
\end{equation}

\begin{equation}
\begin{array}{l}
d_{x_i}^{r_k}=\frac{\overline{d_{x_i}^{r_k}}}{\sum \nolimits_{w = 1}^K {\overline {d_{x_i}^{{r_w}}} }},\\
\overline {d_{x_i}^{{r_k}}} = exp(\frac{{ - {{(k - \gamma )}^2}}}{{2{{\sigma^2} _{r}}}})/{\sigma _{{r}}}, k=1,2,\dots,K
\end{array}
\end{equation}
where $\beta$ and $\gamma$ denote binned ground-truth pitch and roll of the face respectively. Consequently, the training set can be represented as $\{\left ( x_i, \left ( \bm D_i^y,\bm D_i^p,\bm D_i^r \right ) \right ),1\le i \le n\}$ and the goal of the learning becomes to train a set of network parameters $\bm \theta$ to generate a triplet of probability distribution $\left ( \bm F^y \left (x_i;\bm \theta \right ),\bm F^p \left ( x_i;\bm \theta \right ),\bm F^r\left ( x_i;\bm \theta \right ) \right )$ for the three label sets, which is similar to $\left ( \bm D_i^y,\bm D_i^p,\bm D_i^r \right )$.
Wherein,
\begin{equation}
\begin{array}{l}
\bm F^y \left (x_i;\bm \theta \right )=\{f(y_1 |x_i;\bm \theta),f(y_2 |x_i;\bm \theta),\dots,f(y_M |x_i;\bm \theta)\},\\ \sum_{l=1}^{M}f(y_l |x_i;\bm \theta)=1;\\
\bm F^p \left (x_i;\bm \theta \right )=\{f(p_1 |x_i;\bm \theta),f(p_2 |x_i;\bm \theta),\dots,f(p_N |x_i;\bm \theta)\},\\ \sum_{j=1}^{N}f(p_j |x_i;\bm \theta)=1;\\
\bm F^r \left (x_i;\bm \theta \right )=\{f(r_1 |x_i;\bm \theta),f(r_2 |x_i;\bm \theta),\dots,f(r_K |x_i;\bm \theta)\},\\ \sum_{k=1}^{K}f(r_k |x_i;\bm \theta)=1.\\
\end{array}
\end{equation}
The Euclidean distance and Kullback-Leibler (KL) divergence are adopted to construct the loss function measuring the similarity between the ground-truth distribution $\left ( \bm D_i^y,\bm D_i^p,\bm D_i^r \right )$ and predicted distribution $\left ( \bm F^y \left (x_i;\bm \theta \right ),\bm F^p \left ( x_i;\bm \theta \right ),\bm F^r\left ( x_i;\bm \theta \right ) \right )$. The objective of the label distribution learning is to minimize either of the following overall loss functions:

\begin{equation}
\begin{array}{l}
L_{Eu}=\sum \limits_{i=1}^{n} \left\|\bm D_i^y-\bm F^y\left (x_i;\bm \theta \right )\right\|_2 +\sum \limits_{i=1}^{n} \left\|\bm D_i^p-\bm F^p \left ( x_i;\bm \theta \right) \right\|_2\\
+\sum \limits_{i=1}^{n}\left\|\bm D_i^r-\bm F^r\left (x_i;\bm \theta \right)\right\|_2,\\
L_{KL}=\sum \limits_{i=1}^{n}\sum \limits_{l=1}^{M} d_{x_i}^{y_l}\ln \frac{d_{x_i}^{y_l}}{f \left ( y_l | x_i;\bm \theta \right )} +\sum \limits_{i=1}^{n}\sum \limits_{j=1}^{N} d_{x_i}^{p_j} \ln \frac{d_{x_i}^{p_j}}{f\left (p_j| x_i;\bm \theta \right)} + \\
\sum \limits_{i=1}^{n}\sum \limits_{k=1}^{K}d_{x_i}^{r_k} \ln \frac {d_{x_i}^{r_k}}{f\left (r_k| x_i;\bm \theta \right)}
\end{array}
\end{equation}
And we define $L_{GLD}=L_{Eu}+L_{KL}$ as our Gaussian label distribution loss.

\begin{figure*}[htbp]
	\centering
	\includegraphics[width=0.8\linewidth]{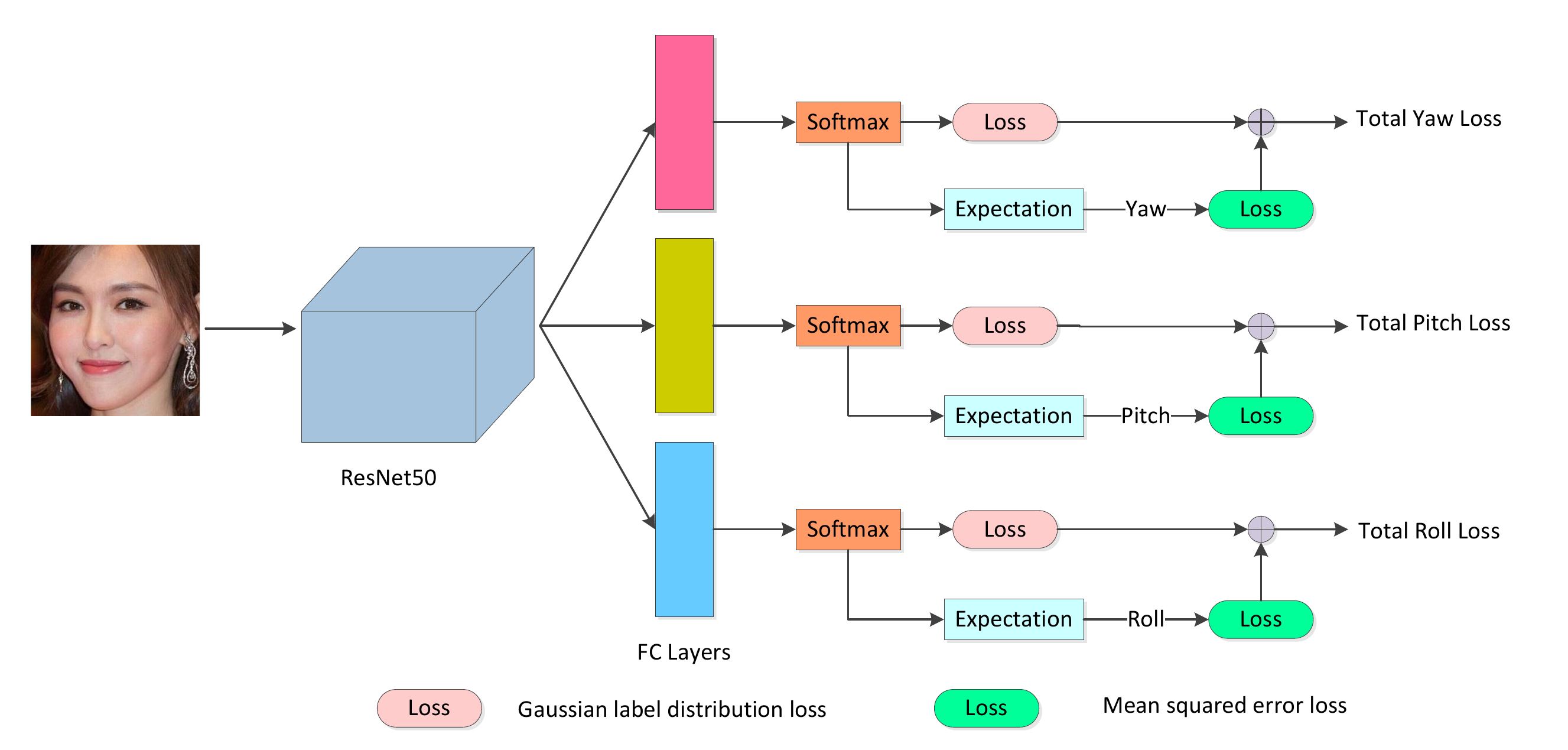}
	\caption{Proposed network architecture for facial pose estimation.}
	\label{fig:3}
\end{figure*}

\subsection{Network Architecture}
We modify the framework presented in Hopenet \cite{16} to construct our network architecture for facial pose estimation. The framework presented in Hopenet \cite{16} originally consists of three separate losses for yaw, pitch and roll respectively and got state-of-the-art result. Each loss is a linear combination of a soft-max cross-entropy loss and a mean squared error(MSE) loss. To achieve better accuracy, we replace the soft-max cross-entropy loss with our Gaussian label distribution loss. Consequently, our learning architecture can be constructed as shown in Fig.\ref{fig:3}.

Our framework consists of a ResNet50 \cite{56}-based backbone network and three branches for yaw, pitch and roll respectively. Each branch is comprised of a fully-connected layer with the number of neurons equal to the total number of corresponding labels and a soft-max layer followed by the combined loss layer. The soft-max operation ensures to satisfy the aforementioned constraints: $\sum_{l=1}^{M}d_{x_i}^{y_l}=1$, $\sum_{j=1}^{N} d_{x_i}^{p_j}=1$ and $\sum_{k=1}^{K} d_{x_i}^{r_k}=1$.

Then the total loss is defined as $L_{total}=L_{GLD}+ \alpha * L_{MSE}$. Wherein, $L_{MSE}$ is the mean squared error loss, and $\alpha$ is a weight used to adjust the two loss components.

\section{Experiments}
\subsection{Training Details}
We choose the 300W-LP \cite{53} and the AFLW \cite{54} to train our network respectively. These two datasets have enough examples with enough different identities and different lighting conditions. The 300W-LP \cite{53}dataset is a collection of popular in-the-wild 2D landmark datasets which have been grouped and re-annotated. The AFLW \cite{54}dataset, which is commonly used to train and test landmark detection methods, also includes pose annotations.

We divide the facial pose into 66 bins within $\pm 99^\circ$ for yaw, pitch and roll respectively, \ie, $M=N=K=66$.  And we set $\sigma_y=\sigma_p=\sigma_r=4$.  All the data is normalized before training by using the ImageNet mean and standard deviation for each color channel. And a pretrained ResNet50\cite{56} on ImageNet is adopted to initialize our network. The proposed multi-loss network is trained with $\alpha=0$, $\alpha=0.01$, $\alpha=0.1$, $\alpha=1$ and $\alpha=2$ on both the 300W-LP dataset and AFLW dataset. All the ten networks are trained using Adam optimization [51] with a learning rate of $10^{-6}$ and $\beta_1=0.9$, $\beta_2=0.999$ and $\varepsilon =10^{-8}$.

\subsection{Results on AFLW2000 and BIWI Benchmark}
The AFLW2000 \cite{53} dataset contains the first 2000 identities of the in-the-wild AFLW \cite{54}dataset with accurate pose annotations. It is an ideal candidate to test our method. The BIWI \cite{53} dataset is collected indoor by recording RGB-D video of different subjects across different facial poses using Kinect v2 device. It is commonly used as benchmark for depth-based pose estimation. Here we will only use the color frames instead of the depth information.

Firstly, we compare our results to the state-of-the-art method Hopenet \cite{16} which is trained using a combination of L2 Euclidean loss and soft-max cross-entropy loss. Then, we compare to the pose estimated from 3DDFA \cite{49} whose primary task is to align facial landmarks, and pose estimated from landmarks using two different landmark detectors: FAN \cite{18} and Dlib\cite{50}, and ground-truth landmarks on both datasets. Additionally, we also list the results of KEPLER \cite{19} on BIWI dataset reported in \cite{16}. Table \ref{tab:1} shows the performance evaluations on AFLW2000 and BIWI Benchmark.

We can see that our best model($\alpha=0.01$) outperforms all other baseline methods by a large margin on AFLW2000 benchmark, reducing the yaw error of the best-performing baselines 3DDFA\cite{49} by 43.9\%, reducing the yaw error of Hopenet\cite{16} by 53.2\%, reducing the pitch error, the roll error, and the mean average error (MAE) of the best-performing baseline Hopenet\cite{16} by 22.8\%, 32.2\%, 36.2\% respectively.

On BIWI benchmark, our method also performs better than all other baseline methods. Our best model($\alpha=0$)trained on 300W-LP dataset reduces the error of the corresponding best-performing baseline Hopenet\cite{16} trained on 300W-LP datatset by 14.3\%, 15\%, 3.7\% and 12.3\% for yaw, pitch, roll and MAE respectively. Our best model($\alpha=0.1$)trained on AFLW dataset also outperforms Hopenet\cite{16} trained on AFLW datatset, reducing the error by 20.8\%, 19.4\%, 0.9\% and 13.8\% for yaw, pitch, roll and MAE respectively.

\begin{table}[htbp]
	\resizebox{\linewidth}{!}{
		\begin{tabular}{|c|c|c|c|c|c|}
			\hline
			Benchmark & Method & Yaw & Pitch & Roll & MAE \\
			\hline
			\multirow{10}{*}{AFLW2000} &Hopenet\cite{16}* & 6.470 & 6.559 & 5.436 & 6.155 \\
			&FAN\cite{18} & 6.358 & 12.277 & 8.714 & 9.116 \\
			&3DDFA\cite{49} & 5.400 & 8.530 & 8.250 & 7.393 \\
			&Dlib\cite{50} & 23.153 & 13.633 & 10.545 & 15.777 \\
			&Ground-truth landmarks & 5.924 & 11.756 & 8.271 & 8.651 \\
			\cline{2-6}
			& Ours($\alpha=0$)* & 3.1791 & 5.3372 & 3.7983 & 4.1049\\
			& Ours($\alpha=0.01$)* & 3.0288 & 5.0634 & 3.6842 & \textbf{3.9255}\\
			&Ours($\alpha=0.1$)* & 3.1446 & 5.2047 & 3.6901 & 4.0131\\
			&Ours($\alpha=1$)* & 3.1064 & 5.3446 & 3.6957 & 4.0489\\
			&Ours($\alpha=2$)* & 3.3236 & 5.3570 & 3.8392 & 4.1733\\
			\hline
			\multirow{17}{*}{BIWI} &Hopenet\cite{16}* & 4.810 & 6.606 & 3.269 & 4.895 \\
			&Hopenet\cite{16}+ & 5.785 & 11.726 & 8.194 & 8.568 \\
			&FAN\cite{18} & 8.532 & 7.483 & 7.631 & 7.882 \\
			&3DDFA\cite{49} & 36.175 & 12.252 & 8.776 & 19.068 \\
			&Dlib\cite{50} & 16.756 & 13.802 & 6.190 & 12.249 \\	
			&KEPLER\cite{19}+ & 8.084 & 17.277 & 16.196 & 13.852 \\	
			\cline{2-6}
			& Ours($\alpha=0$)* & 4.1233 & 5.6142 & 3.1469 & \textbf{4.2948} \\
			& Ours($\alpha=0.01$)* & 4.2367 & 5.8446 & 3.4675 & 4.5163 \\
			&Ours($\alpha=0.1$)* & 4.0967 & 6.0498 & 3.2933 & 4.4799 \\
			&Ours($\alpha=1$)* & 3.9236 & 5.8832 & 3.4014 & 4.4027 \\
			&Ours($\alpha=2$)* & 4.6890 & 6.1271 & 3.3669 & 4.7276 \\
			&Ours($\alpha=0$)+ & 4.5674 & 10.0874 & 8.0633 & 7.5737 \\
			&Ours($\alpha=0.01$)+ & 4.5652 & 8.9595 & 8.7420 & 7.4223 \\
			&Ours($\alpha=0.1$)+ & 4.5839 & 9.4471 & 8.1225 & \textbf{7.3845} \\
			&Ours($\alpha=1$)+ & 4.3564 & 9.2310 & 8.8810 & 7.4895 \\
			&Ours($\alpha=2$)+ & 4.3587 & 9.9015 & 8.6058 & 7.6220 \\
			\hline
		\end{tabular}
	}
	\begin{tablenotes}
		\item[1] \footnotesize *: trained on 300W-LP dataset.
		\item[2] \footnotesize +: trained on AFLW dataset.
	\end{tablenotes}
	\caption{Evaluations on AFLW2000 and BIWI benchmarks.}
	\label{tab:1}
\end{table}

\subsection{Results on AFLW and AFW Benchmark}
\begin{figure}[htbp]
	\centering
	\includegraphics[width=0.8\linewidth]{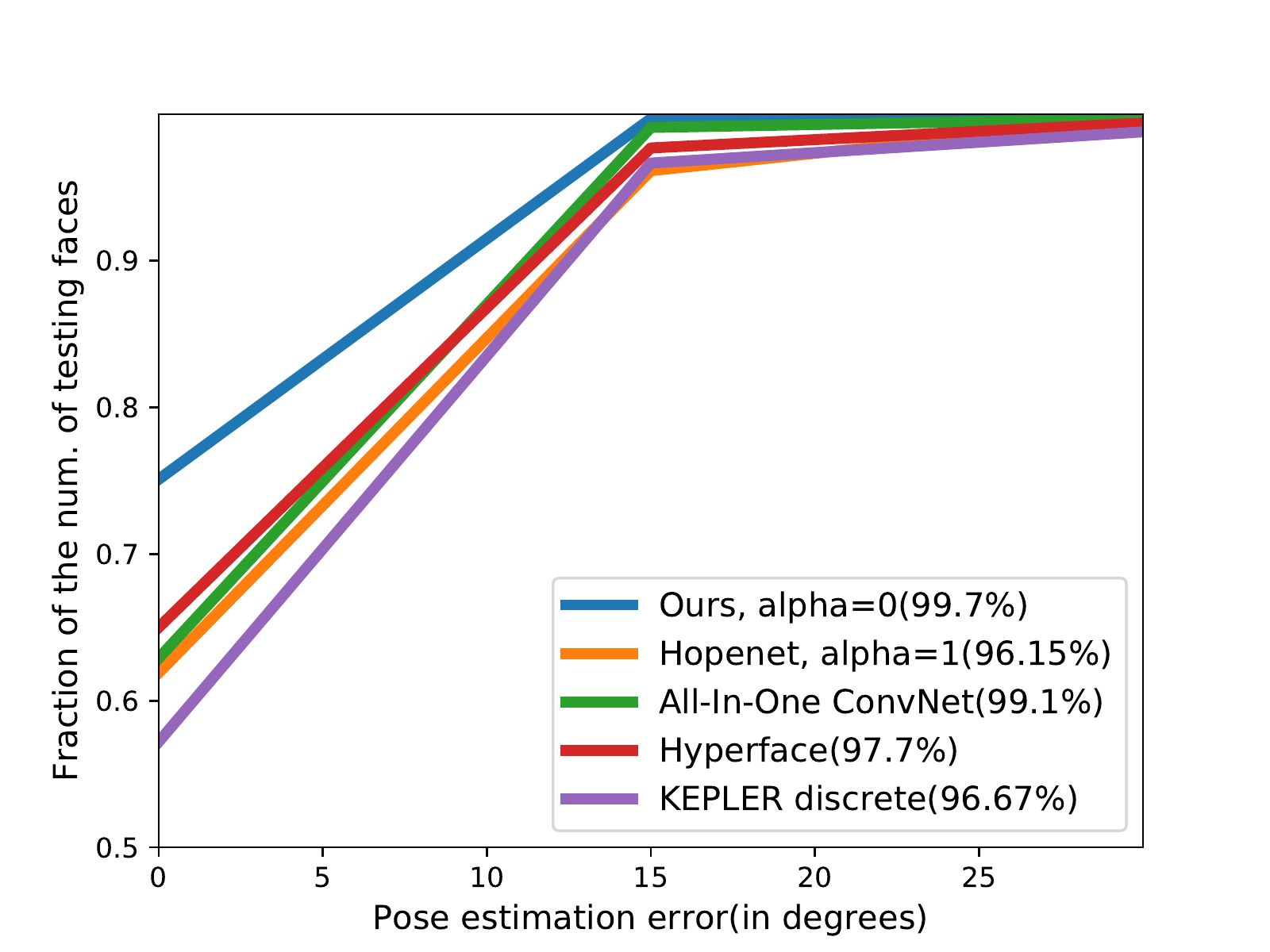}
	\caption{Evaluation on AFW benchmark.}
	\label{fig:5}
\end{figure}

In this section, we present the evaluation results on AFLW \cite{54} and AFW \cite{55} benchmark, using the model trained on AFLW dataset. The AFW \cite{55}benchmark which is commonly used to test landmark detection methods contains rough pose annotations. Here, we compare our results to some deep learning-based methods, including Hopenet \cite{16}, KEPLER \cite{19}, the method proposed by Patacchiola and Cangelosi\cite{17}, Hyperface \cite{21} and All-In-One \cite{20}. Table \ref{tab:3} and Fig.\ref{fig:5} respectively show the results on AFLW and AFW benchmark.

\begin{table}[htbp]
	\centering
	\begin{tabular}{|c|c|c|c|c|}
		\hline
		Method & Yaw & Pitch & Roll & MAE \\
		\hline
		Hopenet\cite{16} & 6.26 & 5.89 &	3.82 & 5.324 \\
		KEPLER\cite{19} & 6.45 & 5.85 & 8.75 & 7.017 \\
		Patacchiola,Cangelosi\cite{17} & 11.04 & 7.15 & 4.4 &7.530 \\		
		\hline
		Ours($\alpha=0$) & 6.83 & 5.26 & 3.92 & 5.34 \\
		Ours($\alpha=0.01$) & 6.00 & 5.31 & 3.75 & \textbf{5.02} \\
		Ours($\alpha=0.1$) & 5.93 & 5.30 & 4.03 & 5.085 \\
		Ours($\alpha=1$) & 5.90 & 5.51 & 3.87 & 5.094 \\
		Ours($\alpha=2$) & 5.90 & 5.62 & 3.77 & 5.097 \\
		\hline
	\end{tabular}
	\caption{Evaluation on AFLW benchmark.}
	\label{tab:3}
\end{table}

We can see that our method outperforms all other baseline methods on AFLW benchmark. Our best model($\alpha=0.01$) reduces the error of the best-performing baseline Hopenet\cite{16} by 4.2\%, 9.85\%, 0.53\% and 5.71\% for yaw, pitch, roll and MAE respectively. On AFW benchmark, our method also performs better than all other baseline methods. Our best model($\alpha=0.01$) achieves a saturated accuracy of over 99\%.

It is noteworthy that, on BIWI and AFLW benchmarks, the improvement of accuracy for roll is much less than for yaw and pitch. We argue that two reasons result in this situation. One reason is that, the distribution of training sets in roll domain is extremely imbalanced compared to that in yaw and pitch domains(as shown in Fig.\ref{fig:1}), and the most of training examples lie in the area of small roll, which limits the learning ability of our method in roll domain, especially in the area of large roll. The other reason is that the test sets also have the similar characteristic as the first reason mentioned. In test sets, 67.65\% examples of BIWI and 65.57\% examples of AFLW lie in $\pm 10^\circ$ for roll, while 33.54\% of BIWI and 26.23\% of AFLW for yaw, and 22.97\% of BIWI and 47.13\% of AFLW for pitch. That is, the BIWI and AFLW benchmarks have relatively few examples with large roll. Both reasons restrict the improvement our method can make for roll.
\section{Conclusion}

This paper presents a novel computational model for facial pose estimation, which is reformulated as label distribution learning problem rather than the conventional single-label supervised learning. This makes a face image contribute to not only the learning of its real pose, but also the learning of its adjacent poses, mitigating the degradation of pose predictor caused by the lack of sufficient training data. Experiments on several popular benchmarks show our method is state-of-the-art.

{\small
\bibliographystyle{ieee}
\bibliography{egbib}
}

\end{document}